\pdfoutput=1
\documentclass{ecai}
\usepackage{times}
\usepackage{graphicx}
\usepackage{latexsym}

\usepackage{epsfig}
\usepackage{amsthm,amsmath,amsfonts,amssymb}
\usepackage{algorithm,algpseudocode,amsmath}
\usepackage{url}
\usepackage{subcaption}
\usepackage{multirow}

\newcommand{\argmax}{\mathrm{argmax}}

\makeatletter

\let\oldReturn\Return
\renewcommand{\Return}{\State\oldReturn}

\ecaisubmission 

\begin{document}

\title{Event Recognition with Automatic Album Detection based on Sequential Processing, Neural Attention and Image Captioning}

\author{Andrey V. Savchenko\institute{National Research University Higher School of Economics,
Laboratory of Algorithms and Technologies for Network Analysis, Nizhny Novgorod, Russia, email: avsavchenko@hse.ru} \ \institute{St. Petersburg Department of Steklov Institute of Mathematics,
Samsung-PDMI Joint AI Center, St. Petersburg, Russia} }

\maketitle
\bibliographystyle{ecai}

\begin{abstract}
In this paper a new formulation of event recognition task is examined: it is required to predict event categories in a gallery of images, for which albums (groups of photos corresponding to a single event) are unknown. We propose the novel two-stage approach. At first, features are extracted in each photo using the pre-trained convolutional neural network. These features are classified individually. The scores of the classifier are used to group sequential photos into several clusters. Finally, the features of photos in each group are aggregated into a single descriptor using neural attention mechanism. This algorithm is optionally extended to improve the accuracy for classification of each image in an album. In contrast to conventional fine-tuning of convolutional neural networks (CNN) we proposed to use image captioning, i.e., generative model that converts images to textual descriptions. They are one-hot encoded and summarized into sparse feature vector suitable for learning of arbitrary classifier. Experimental study with Photo Event Collection and Multi-Label Curation of Flickr Events Dataset demonstrates that our approach is 9-20\% more accurate than event recognition on single photos. Moreover, proposed method has 13-16\% lower error rate than classification of groups of photos obtained with hierarchical clustering. It is experimentally shown that the image captions trained on Conceptual Captions dataset can be classified more accurately than the features from object detector, though they both are obviously not as rich as the CNN-based features. However, it is possible to combine our approach with conventional CNNs in an ensemble to provide the state-of-the-art results for several event datasets.
\end{abstract}

\section{INTRODUCTION}
People are taking more photos than ever before in recent years~\cite{guo2017multigranular} due to the rapid growth of social networks, cloud services and mobile technologies. To organize a personal collection, the photos are usually assigned to albums according to some events. The photo organizing systems (Apple iPhoto, Google Photos, etc.) allow the user to rapidly search for required photo, and also to increase the efficiency of work with a gallery~\cite{sokolova2017organizing}. Nowadays, these systems usually include content-based image analysis and automatic association of each photo with different tags (scene description, persons, objects, locations, etc.). Such analysis can be used not only to selectively retrieve photos for particular tag in order to keep nice memories of some episodes of user's live~\cite{wang2018transferring}, but to make personalized recommendations that assist customers in finding relevant items within large collections. The design of such systems requires the careful consideration of the user modeling approach~\cite{savchenko2019user}. A large gallery of photos on a mobile device can be used for understanding of such user's interests as sport, gadgets, fitness, cloth, cars, food, travelling, pets, etc.~\cite{grechikhin2019user,rassadin2019scene}. 

In this paper we focus on one of the most challenging parts of photo organizing engine, namely, image-based event recognition~\cite{ahmad2019deep}, in order to extract such events as holidays, sport events, weddings, various activities, etc. An event can be defined as a category that captures the ``complex behavior of a group of people, interacting with multiple objects, and taking place in a specific environment"~\cite{wang2018transferring}. There exist two different tasks of event recognition. The first one is focused on processing of single photos, i.e. event is considered as a complex scene with large variations in visual appearance and structure~\cite{wang2018transferring}. The second task aims at predicting the event
categories of a group of photos (album)~\cite{bacha2016event}. In the latter case it is assumed that all photos in an album are weakly labeled~\cite{ahmad2017event}, though importance of each image may differ~\cite{wang2017recognizing}. However, in practice only a gallery of photos is available so that the latter approach requires a user to manually choose the albums. Another option includes location-based album creation if the GPS tags are switched on. In both cases the usage of album-based event recognition is limited or even impossible.

Thus, in this paper we consider the new task of event recognition, in which a gallery of photos is given and it is known that it contains ordered albums with unknown borders. We propose to automatically assign these borders based on the visual content of consecutive photos in a gallery. Next, consecutive photos are grouped, and descriptor of each group is computed with an attention mechanism from the neural aggregation module~\cite{yang2017neural}. Finally, this approach is extended as follows. Despite conventional usage of CNNs as discriminative models in a classifier design, we propose to borrow generative models to represent an input image in other domain. In particular, we use existing methods of image captioning~\cite{hossain2019comprehensive} that generates textual descriptions of images. Our main contribution is a demonstration that the generated descriptions can be fed to the input of classifier in an ensemble in order to improve event recognition accuracy of traditional methods. Though the proposed visual representation is not as rich as features extracted by fine-tuned CNNs, they are better than the outputs of object detectors~\cite{rassadin2019scene}. 

\section{LITERATURE SURVEY} \label{sec:2}

Annotating personal photo albums is an emerging trend in photo organizing services~\cite{dao2011signature}. A method for hierarchical photo organization into topics and topic-related categories on a smartphone is proposed in~\cite{lonn2019smartphone} based on integration of convolutional neural network (CNN) and topic modeling for image classification. An automatic hierarchical clustering and best photo selection solution is introduced in~\cite{kuzovkin2019context} for modeling user decisions in organizing or clustering similar photos in albums. Organizing photo albums for user preference prediction on mobile device is considered in~\cite{savchenko2019efficient}. 

The task of event recognition in the personal photo collections is not to recognize event in individual photo but in the whole album.
The events and sub-events of the sub-sequence photos are identified in~\cite{dao2011signature} by integrating the optimized linear programming with the color descriptor of the signature image. The Stopwatch Hidden Markov Models were applied in~\cite{bossard2013event} by treating the photos in an album as sequential data. 
The paper~\cite{ahmad2017event} tackles the presence of irrelevant images in an album with multiple instance learning techniques. An iterative updating procedure for event type and image importance score prediction in a siamese network is presented in~\cite{wang2017recognizing}. The authors of this paper used a CNN that recognizes the event type, and a Long Short-Term Memory (LSTM)-based sequence level event recognizer in a whole album. Moreover, they successfully applied the method for learning representative deep features for image set analysis~\cite{wu2015learning}. The latter approach focuses on capturing the co-occurrences and frequencies of features so that the temporal coherence of photos in an album is not required. A model to recognize events from coarse to fine hierarchical level using multi-granular features~\cite{savchenko2019efficient} is proposed in~\cite{guo2017multigranular} based on an attention network that learns the representations of photo albums. The efficiency of re-finding expected photos in mobile phones was improved by a method to classify personal photos based on relationship of shooting time and shooting location to specific events~\cite{geng2018classifying}.

The album information is not always available so that a gallery contains unstructured list of photos ordered by their creation time. In such case it is possible to use existing methods of event recognition on single photos~\cite{ahmad2019deep}. Similar to other computer vision domains, the mainstream approach tends to applications of CNN-based architectures. For example, four different layers of fine-tuned CNN were used to extract features and perform Linear Discriminant Analysis in order to obtain the top entry in the ChaLearn LAP 2015 cultural event recognition challenge~\cite{escalera2015chalearn}. The bounding boxes of detected objects are projected onto multi-scale spatial maps for increasing the accuracy of event recognition~\cite{xiong2015recognize}. The novel iterative selection method is introduced in~\cite{wang2018transferring} to identify a subset of classes that are most relevant for transferring deep representations learned from object (ImageNet) and scene (Places2) datasets. 

Unfortunately, the accuracy of event classification on still photos~\cite{wang2018transferring} is in general much lower than the accuracy of album-based recognition~\cite{wang2017recognizing}. That is why in this paper we proposed to concentrate on other suitable visual features extracted with the generative models and, in particular, image captioning techniques. There is a wide range of applications of image captioning: from automatic generation of descriptions for photos posted in social networks to image retrieval from databases using generated text descriptions~\cite{vijayaraju2019image}. The image captioning methods are usually based on an encoder-decoder neural network, which first encodes an image into a fixed-length vector representation using pre-trained CNN, and then decodes the representation into captions (a natural language description). During the training of a decoder (generator) the input image and its ground-truth textual description are fed as inputs to the neural network, while one hot encoded description presents the desired network output. Description is encoded using text embeddings in the Embedding (look-up) layer~\cite{goodfellow2016deep}. The generated image and text embeddings are merged using concatenation or summation, and form the input to the decoder part of the network. It is typical to include the recurrent neural network (RNN) layer followed by a fully connected layer with the Softmax output layer.

One of the first successful models, ``Show and Tell"~\cite{cap_ST}, won the first MS COCO Image Captioning Challenge in 2015. It uses the RNN with the long short-term memory (LSTM) units in a decoder part.  Its enhancement ``Show, Attend and Tell"~\cite{cap_SAT} incorporates a soft attention mechanism to improve the quality of the caption generation. The ``Neural Baby Talk" image captioning model~\cite{cap_NBT} is based on generating of the template with slot locations explicitly tied to specific image regions. These slots are then filled in by visual concepts identified in the object detectors. The foreground regions are obtained using the Faster-RCNN network~\cite{ren2015faster}, and the LSTM with attention mechanism serves as a decoder. The ``Multimodal Recurrent Neural Network" (mRNN)~\cite{cap_mRNN} is based on the Inception network for image features extraction and deep RNN for sentences generation. One of the best models nowadays is the Auto-Reconstructor Network (ARNet)~\cite{cap_ARNet}, which uses Inception-V4 network~\cite{cap_inc4} in an encoder and the decoder is based on LSTM. There exist two pre-trained models with greedy search (ARNet-g) and beam search (ARNet-b) with size 3 to generate the final caption for each input image.

\section{MATERIALS AND METHODS}\label{sec:3}

\subsection{Problem formulation}\label{sec:3.1}
As it was noticed above, an important task is automatic extraction of albums from a personal gallery based on a visual content of photos. In this subsection we discover a technological engine that can solve this task by using sequential processing of photos similarly to cluster analysis with the Basic Sequential Algorithmic Scheme (BSAS)~\cite{ashour2019improve}. Our main task can be formulated as follows. It is required to assign each photo $X_t, t \in \{1,...,T\}$ from a gallery of an input user to one of $C>1$ event categories (classes). Here $T \ge 1$ is the total number of photos in a gallery. The training set of $N \ge 1$ albums is available for learning of event classifier. The $n$-th reference album is defined by $L_n$ images $\{X_n(1),...,X_n(L_n)\}$. The class label $c_n\in \{1,...,C\}$ of each $n$-th album is supposed to be given, i.e., we assume that an album is associated with exactly one event type. 

Conventional event recognition on single photos~\cite{wang2018transferring} is the special case of above-formulated problem if $T=1$. The main difference is the following assumption. The gallery $\{X_t\}$ is not a random collection of photos but can be represented as a \textit{sequence of disjoint albums}. Each image in an album is associated with the same event. In contrast to the album-based event recognition, the borders of each album are unknown. This task possesses several characteristics that makes it extremely challenging compared to previously studied problems. One of these characteristics is the presence of irrelevant images or unimportant photos that can be in principle associated to any event~\cite{ahmad2019deep}. These images are easily detected in attention-based models~\cite{guo2017multigranular,yang2017neural}, but may have a significant impact on a quality of automatic album selection.

The baseline approach here is to classify all $T$ photos independently. In such case it is typical to unfold the training albums into a set $\mathbf{X}=\{X_1(1),...,X_1(L_1), X_2(1),...,X_2(L_2),...,X_N(1),...,X_N(L_N)\}$ of $L=L_1+...+L_N$ photos so that the collection-level label $c_n$ of the $n$-th album is assigned to labels of each $l$-th photo ($l \in \{1,...,L_n\}$). Next, it is possible to train an arbitrary event classifier. If $L$ is rather small to train a deep CNN from scratch, the transfer learning or domain adaptation can be applied~\cite{goodfellow2016deep}. In these methods a large external dataset, e.g. ImageNet-1000 or Places2~\cite{zhou2018places}, is used to pre-train a deep CNN. As we pay special attention to offline recognition on mobile devices, it is reasonable to use such CNNs as MobileNet v1/v2~\cite{howard2017MobileNets,sandler_inverted_2018}. The final step in transfer learning is fine-tuning of this neural network on $\mathbf{X}$. This step includes replacement of the last layer of the pre-trained CNN to the new layer with Softmax activations and $C$ outputs. During the classification process, each input image $X_t$ is fed to the fine-tuned CNN to compute the scores (predictions at the last layer):

\begin{equation*}
\mathbf{p}_t=[p_{1;t},...,p_{C;t}], \sum \limits_{c=1}^{C}{p_{c;t}}=1.
\end{equation*}

This procedure can be modified by replacing $C$ logistic regressions in the last layer to more complex classifier, e.g., random forest (RF), support vector machine (SVM) or gradient boosting. In this case the features (embeddings)~\cite{savchenko2019sequential} are extracted using the outputs of one of the last layers of pre-trained CNN. Namely, the images $X_t$ and $X_n(l)$ are fed to the CNN, and the outputs of the one-but-last layer are used as the $D$-dimensional feature vectors $\mathbf{x}_t=[x_{1;t},...,x_{D;t}]$ and $\mathbf{x}_n(l)=[x_{n;1}(l),...,x_{n;D}(l)]$, respectively. Such deep learning-based feature extractors allow training of a general classifier $\mathcal{C}$. The $t$-th photo is fed into this classifier to obtain $C$-dimensional confidence scores $\mathbf{p}_t$.

Finally, the confidences $\mathbf{p}_t$ computed by any of above-mentioned ways are used to make a decision in favor of the most probable class:

\begin{equation}
c^*(t)=\underset{c \in \{1,...,C\}}{\operatorname{argmax}}\ p_{c;t}.
\end{equation}

\subsection{Event recognition in a gallery of photos}
\begin{figure}
 \centering
 \includegraphics[width=1\linewidth]{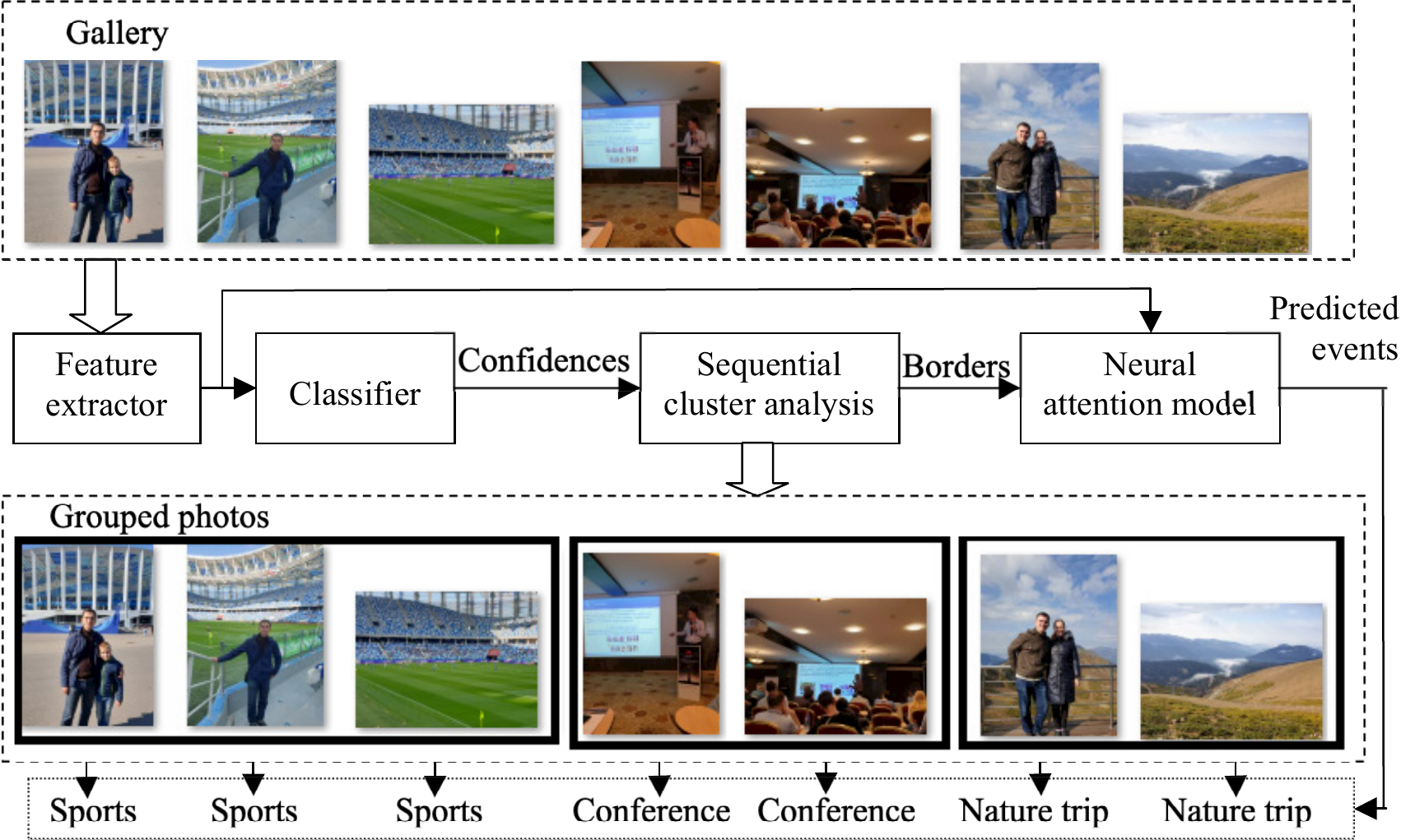}
 \caption{Proposed gallery-based event recognition pipeline}
\label{fig:final_pipeline}
\end{figure}

The proposed pipeline is presented in Fig.~\ref{fig:final_pipeline}. 

Here, firstly, module ``Feature extractor" computes embeddings $\mathbf{x}_t$ of every $t$-th individual photo as described in previous Subsection~\ref{sec:3.1}. The classifier confidences $\mathbf{p}_t$ are estimated in the ``Classifier" block. Next, we use sequential analysis from BSAS clustering~\cite{ashour2019improve} in the ``Sequential cluster analysis" module for a sequence of \textit{confidences} $\{\mathbf{p}_t\}$ in order to obtain the borders of albums. Namely, the distances between neighbor photos $\rho(\mathbf{p}_t,\mathbf{p}_{t-1})$ are computed. If a distance does not exceed a certain threshold $\rho_0$ then it is assumed that both photos are included in the same album. If location information is available in EXIF (Exchangeable Image File Format) data of these photos, the distance between their locations can be added to $\rho(\mathbf{p}_t,\mathbf{p}_{t-1})$ in order to obtain the final distance to be matched with a threshold. Otherwise, the border between two albums is established at the $t$-th position. As a result, we obtain the borders $1 \le t_1<...< t_K=T$ of $K \in \{1,...,T\}$ albums, so that the $k$-th album contains photos $X(t), t \in \{t_{k-1}+1,...,t_k\}$, where $t_0=0$.

\begin{figure}
 \centering
 \includegraphics[width=1\linewidth]{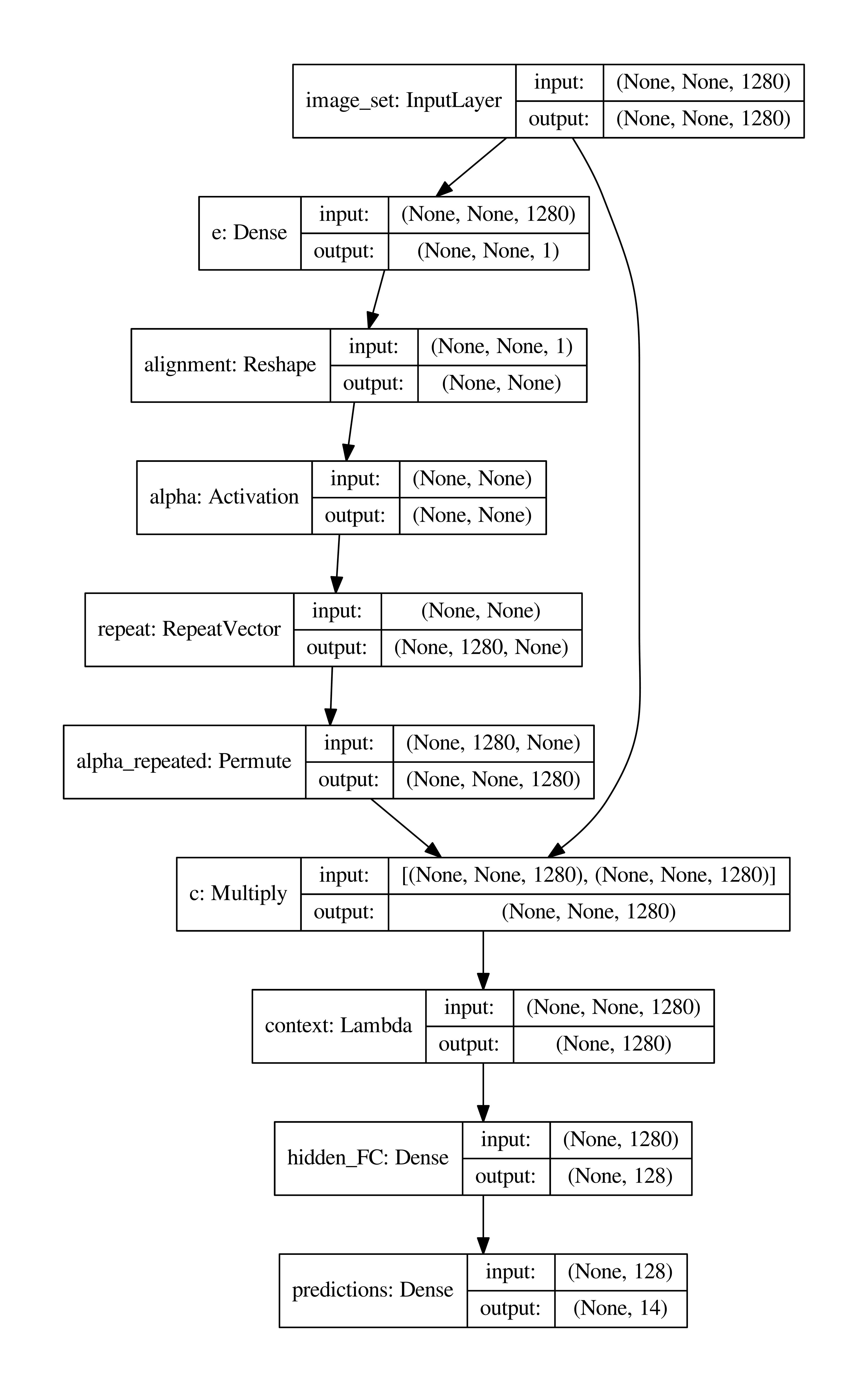}
 \caption{Attention-based neural network for embeddings from MobileNet v2}
\label{fig:network}
\end{figure}

At the second stage, the final descriptor $\mathbf{x}(k)$ of the $k$-th album is produced as a weighted sum of individual  features $\mathbf{x}_t$: 

\begin{equation}\label{eq:weighted_sum}
 \mathbf{x}(k)=\sum \limits_{t=t_{k-1}+1}^{t_k}{w(\mathbf{x}_t)\mathbf{x}_t},
\end{equation}
where the weights $w$ may depend on the features $\mathbf{x}_t$. It is typical to use here average pooling (AvgPool) with equal weights, so that conventional computation of mean feature vector is implemented. However, in this paper we propose to learn the weights $w(\mathbf{x}_t)$, particularly, with an attention mechanism from the neural aggregation module used previously only for video recognition~\cite{yang2017neural}:

\begin{equation}
\label{eq:attention_1}
 w(\mathbf{x}_t) = \frac{\exp(\mathbf{q}^T\mathbf{x}_t)}{\sum \limits_{j=t_{k-1}+1}^{t_k}{\exp(\mathbf{q}^T\mathbf{x}_j)}}.
\end{equation}

Here $\mathbf{q}$ is the learnable $D$-dimensional vector of weights. The dense (fully connected) layer is attached to the resulted descriptor $\mathbf{x}(k)$, and the whole neural network (Fig.~\ref{fig:network}) is trained in end-to-end manner using given training set of $N \ge 1$ albums. The event class predicted by this network in the ``Neural attention model" block (Fig.~\ref{fig:final_pipeline}) is assigned to all photos $X(t), t \in \{t_{k-1}+1,...,t_k\}$.

\begin{algorithm}[t!]
\caption{Proposed gallery-based event recognition} 
\label{algorithm1}
\begin{algorithmic}[1]
\Require input gallery $X(t), t \in \{1,...,T\}$, classifier $\mathcal{C}$, threshold $\rho_0$
\Ensure event labels $c^*(t)$ of all input images
\State Assign $K:=0$, initialize list of borders $B:=[]$
 \For{each input image $t \in \{1,...,T\}$}
 \State Feed the $t$-th image into a CNN and compute embeddings $\mathbf{x}_t$
 \State Compute confidences $\mathbf{p}_t$ using classifier $\mathcal{C}$
 \If {$t=1$ or $\rho(\mathbf{x}_t,\mathbf{x}_{t-1})>\rho_0$}
\State Assign $K:=K+1$, append $t-1$ to the list $B$
 \EndIf
 \EndFor
\State Append $T$ to the list $B$
 \For{each extracted album $k \in \{1,...,K\}$}
\State Feed input images $\{X_{B[k-1]+1},X_{B[k-1]+2},...,X_{B[k]}\}$ into attention network (\ref{eq:weighted_sum})-(\ref{eq:attention_1}) and obtain event class $c^*$
\State Assign $c^*(t):=c^*$ for all $t\in \{B[k-1]+1,...,B[k]\}$
 \EndFor
 \Return labels $c^*(t),t \in \{1,...,T\}$
 \end{algorithmic}
\end{algorithm}

\begin{algorithm}[t!]
\caption{Learning procedure in the proposed approach} 
\label{algorithm2}
\begin{algorithmic}[1]
 \For{each album $n \in \{1,...,N\}$}
 \For{each image $l \in \{1,...,L_n\}$}
 \State Feed image $X_n(l)$ into a CNN and compute embeddings $\mathbf{x}_n(l)$
 \EndFor
 \EndFor
 \State Train classifier $\mathcal{C}$ using unfolded training set $\mathbf{X}$ of embeddings
 \State Train attention network (\ref{eq:weighted_sum})-(\ref{eq:attention_1}) using subsets with fixed size $S$ of all training sets of features $\{\mathbf{x}_n(l)\}$
 \For{each album $n \in \{1,...,N\}$}
 \For{each image $l \in \{1,...,L_n\}$}
 \State Predict confidence scores $\mathbf{p}_n(l)$ for embeddings $\mathbf{x}_n(l)$ using classifier $\mathcal{C}$
 \EndFor
 \EndFor
 \State Randomly permute all indices $\{1,...,N\}$ to obtain sequence $(n_1,...,n_N)$
 \State Unfold all training embeddings using this permutation: $\mathbf{\tilde{X}}=\{X_{n_1}(1),...,X_1(L_{n_1}),...,X_{n_N}(1),...,X_{n_N}(L_{n_N})\}$
 \State Assign $\rho:=0, \alpha^*:=0$
 \For{each potential threshold $\rho$}
 \State Call Algorithm~\ref{algorithm1} with parameters $\mathbf{\tilde{X}}, \mathcal{C}$ and threshold $\rho$
 \State Compute accuracy $\alpha$ using predictions for all training images
 \If {$\alpha^*<\alpha$} 
 \State Assign $\alpha^*:=\alpha, \rho_0:=\rho$
 \EndIf
 \EndFor
 \Return classifier $\mathcal{C}$, attention network, threshold $\rho_0$
 \end{algorithmic}
\end{algorithm}

Complete classification and learning procedures are presented in Algorithm~\ref{algorithm1} and Algorithm~\ref{algorithm2}, respectively. For simplicity, we mentioned that the latter calls the event prediction in step 17. However, to speed-up computations it is recommended to pre-compute the pair-wise distance matrix between confidence scores of all training images so that feature extraction (steps 3-4 in Algorithm~\ref{algorithm1}) and distance calculation are not needed during the learning of our model.

\begin{figure*}
 \centering
\begin{subfigure}{.3\textwidth}
 \centering
 \includegraphics[height=8.5cm]{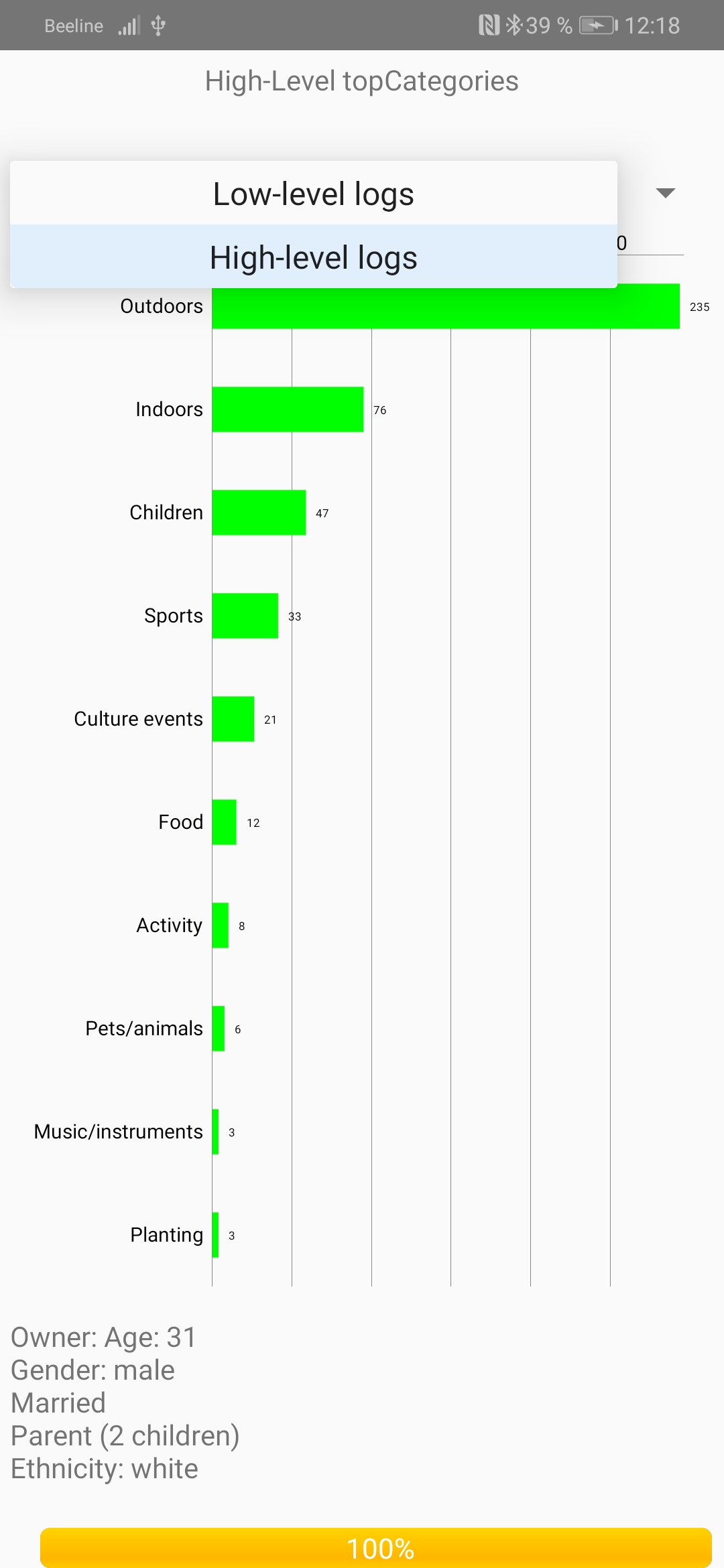}
 \caption{}
 \label{fig:demoUI_sfig1}
\end{subfigure}
\begin{subfigure}{.3\textwidth}
 \centering
 \includegraphics[height=8.5cm]{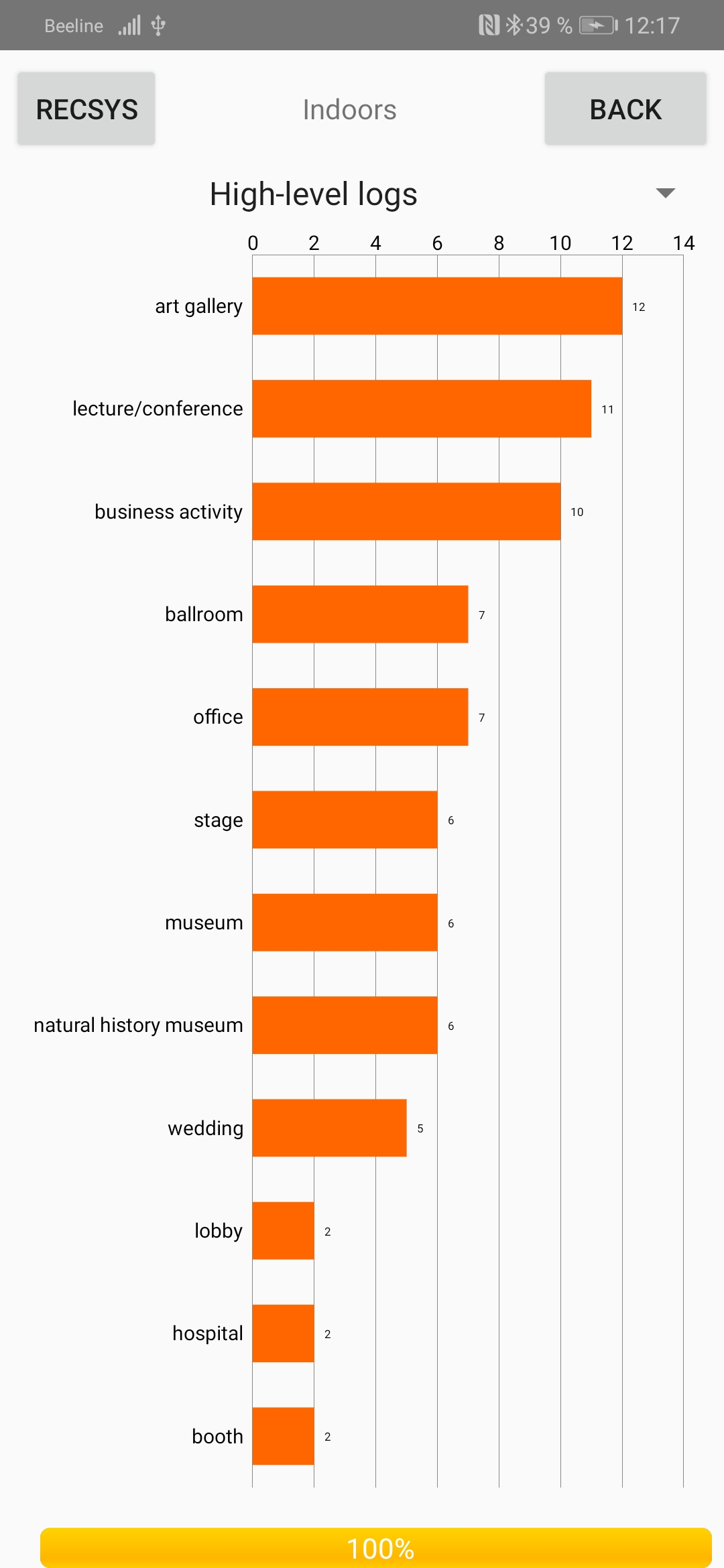}
 \caption{}
 \label{fig:demoUI_sfig2}
\end{subfigure}
\begin{subfigure}{.3\textwidth}
 \centering
 \includegraphics[height=8.5cm]{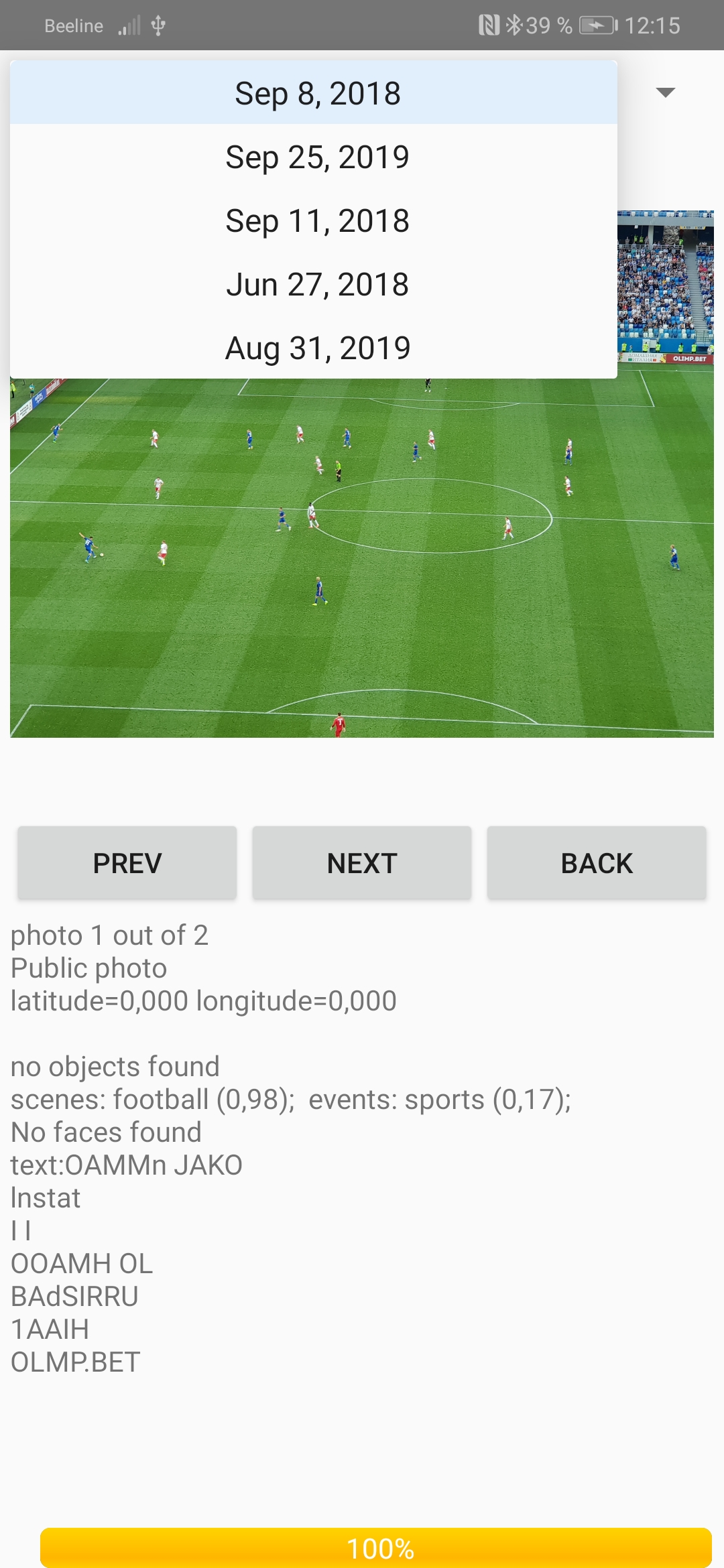}
 \caption{}
 \label{fig:demoUI_sfig3}
\end{subfigure}
 \caption{Mobile demo GUI}
\label{fig:mobile_demo_UI}
\end{figure*}

We implemented the whole pipeline (Fig.~\ref{fig:final_pipeline}) in the publicly-available demo application for Android\footnote{\url{https://drive.google.com/open?id=1aYN0ZwU90T8ZruacvND01hbIaJS4EZLI}} (Fig.~\ref{fig:mobile_demo_UI}), that was previously developed to extract user preferences by processing all photos from the gallery in the background thread~\cite{savchenko2019user}. The similar events found in photos made in one day were united into High-level logs for the most important events. We display only those scenes/events for which there exist at least 2 photos and the average score of scene/event predictions for all photos of the day exceeds a certain threshold. The sample screenshot of the main user interface is shown in Fig.~\ref{fig:demoUI_sfig1}. It is possible to tap any bar in this histogram to show a new form with detailed categories (Fig.~\ref{fig:demoUI_sfig2}). If a concrete category is tapped, a ``display'' form appears, which contains a list of all photos from the gallery with this category (Fig.~\ref{fig:demoUI_sfig3}). Here we group event by date and provide a possibility to choose concrete day.

\subsection{Event recognition in single photos}
Event recognition in single photos task can be formulated as a typical image recognition problem. It is required to assign an input photo $X$ from a gallery to one of $C>1$ event categories (classes). The training set of $N \ge 1$ images $\mathbf{X}=\{X_n| n \in \{1,...,N\} \}$ with known event labels $c_n\in \{1,...,C\}$ is available for classifier learning. Sometimes the training photos of the same event are associated with an album~\cite{bossard2013event,wang2017recognizing}. In such case the training albums are unfolded into a set $\mathbf{X}$ so that the collection-level label of the album is assigned to labels of each photo from this album. This task possesses several characteristics that makes it extremely challenging compared to album-based event recognition. One of these characteristics is the presence of irrelevant images or unimportant photos that can be associated to any event~\cite{ahmad2019deep}. These images can be  detected by attention-based models when the whole album is available~\cite{guo2017multigranular}, but may have a significant impact on a quality of event recognition in single images.

As $N$ is usually rather small, the transfer learning may be applied~\cite{goodfellow2016deep}. A deep CNN is firstly pre-trained on a large dataset, e.g. ImageNet or Places~\cite{zhou2018places}. Secondly, this CNN is fine-tuned on $\mathbf{X}$, i.e., the last layer is replaced to the new layer with Softmax activations and $C$ outputs. An input image $X$ is classified by feeding it to the fine-tuned CNN to compute $C$ scores from the output layer, i.e., estimates of posterior probabilities for all event categories. This procedure can be modified by extraction of deep image features (embeddings) using the outputs of one of the last layers of the pre-trained CNN. The images $X$ and $X_n$ are fed to the input of the CNN, and the outputs of the one-but-last layer are used as the $D$-dimensional feature vectors $\mathbf{x}=[x_1,...,x_D]$ and $\mathbf{x}_n=[x_{n;1},...,x_{n;D}]$, respectively. Such deep learning-based feature extractors allow training of a general classifier $\mathcal{C}_{emb}$, e.g., k-nearest neighbor, random forest (RF), support vector machine (SVM) or gradient boosting. The $C$-dimensional vector of $\mathbf{p}_{emb}=\mathcal{C}_{emb}(\mathbf{x})$ confidence scores is predicted given the input image in both cases of fine-tuning with the last Softmax layer in a role of classifier $\mathcal{C}_{emb}$ and feature extraction with general classifier. The final decision can be made in favor of class with the maximal confidence.

\begin{figure}
	\includegraphics[width=1\linewidth]{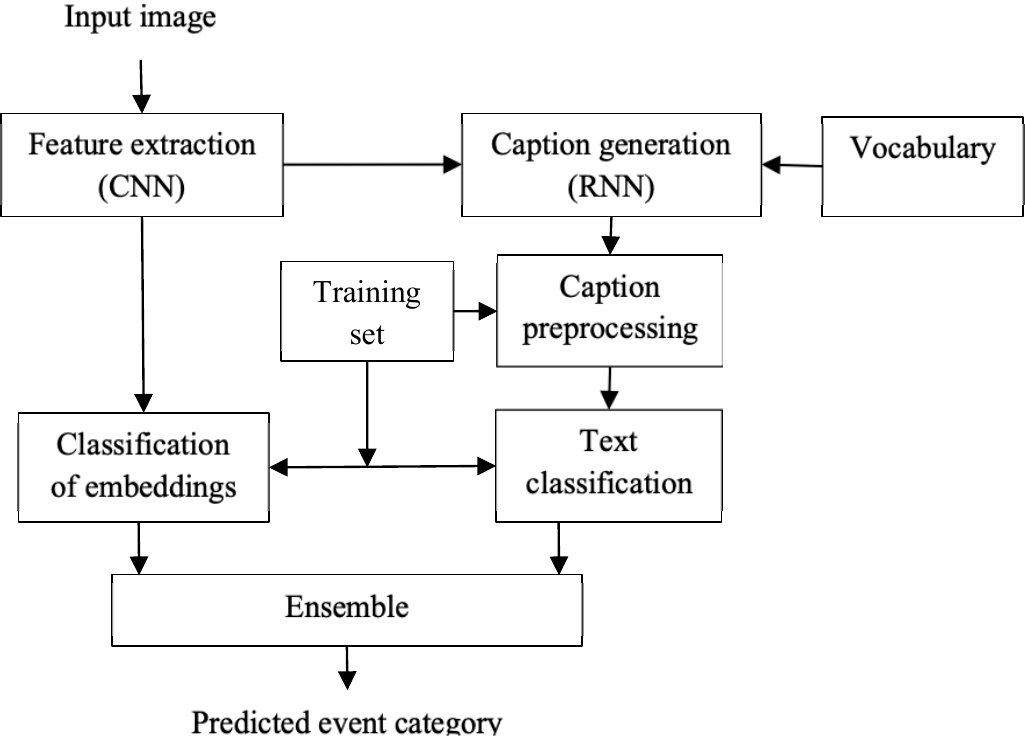}	
	\centering
	\caption{Proposed event recognition pipeline based on image captioning } 
	\label{cap_pipeline}
\end{figure}

In this paper we use another approach to event recognition based on generative models and image captioning. The proposed pipeline is presented in Fig.~\ref{cap_pipeline}. At first, conventional extraction of embeddings $\mathbf{x}$ is implemented using pre-trained CNN. Next, these visual features and a vocabulary $V$ are fed to a special RNN-based neural network (generator) that produces the caption, which describes the input image. Caption is represented as a sequence of $L>0$ tokens $\mathbf{t}=\{t_0, t_1...,t_{L+1}\}$ from the vocabulary ($t_l \in V, l\in \{0,...,L\}$). It is generated sequentially, word-by word starting from $t_0=<START>$ token until a special $t_L=<END>$ word is produced~\cite{cap_ARNet}. 

The generated caption $\mathbf{t}$ is fed into an event classifier. In order to learn its parameters, every $n$-th image from the training set is fed to the same image captioning network to produce the caption $\mathbf{t}_n=\{t_{n;0}, t_{n;1}...,t_{n;L_n+1}\}$. Since the number of tokens $L_n$ is not the same for all images, it is necessary to either train a sequential RNN-based classifier or transform all captions into a feature vectors with the same dimensionality. As the number of training instances $N$ is not very large, we experimentally noticed that the latter approach is as accurate as the former, though the training time is significantly lower. Hence, we decided to use the one-hot encoding of the sequences $\mathbf{t}$ and $\{\mathbf{t}_n\}$ into vectors of 0s and 1s as described in~\cite{francois2017deep}. In particular, we select a subset of vocabulary $\tilde{V} \subset V$ by choosing top most frequently occuring words in the training data $\{\mathbf{t}_n\}$ with optional exclusion of stop words. Next, the input image is represented as the $|\tilde{V}|$-dimensional sparse vector $\mathbf{\tilde{t}} \subset \{0,1\}^{|\tilde{V}|}$, where $|\tilde{V}|$ is the size of reduced vocabulary $\tilde{V}$ and the $v$-th component of vector $\mathbf{\tilde{t}}$ is equal to 1 only if at least one of $L$ words in the caption $\mathbf{t}$ is equal to the $v$-th word from vocabulary $\tilde{V}$. This would mean, for instance, turning the sequence \{1, 5, 10, 2\} into a $\tilde{V}$-dimensional sparse vector that would be all 0s except for indices 1, 2, 5 and 10, which would be 1s~\cite{francois2017deep}. The same procedure is used to describe each $n$-th training image with $\tilde{V}$-dimensional sparse vector $\mathbf{\tilde{t}}_n$. After that an arbitrary classifier $\mathcal{C}_{txt}$ of such textual representations suitable for sparse data can be used to predict $C$ confidence scores $\mathbf{p}_{txt}=\mathcal{C}_{txt}(\mathbf{\tilde{t}})$. It was demonstrated in~\cite{francois2017deep} that such approach is even more accurate than conventional RNN-based classifiers (including one layer of LSTMs) for IMDB dataset.

\begin{figure*}[h!]
 \centering
\begin{subfigure}{0.49\textwidth}
 \centering
 \includegraphics[height=6cm]{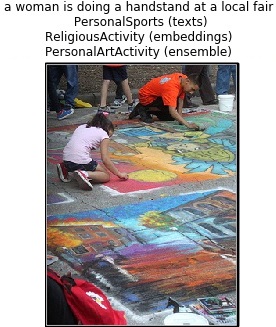}
 \caption{}
 \end{subfigure}
\begin{subfigure}{0.49\textwidth}
 \centering
 \includegraphics[height=6cm]{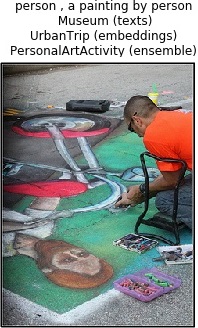}
 \caption{}
 \end{subfigure}
 
\begin{subfigure}{0.49\textwidth}
 \centering
 \includegraphics[height=6cm]{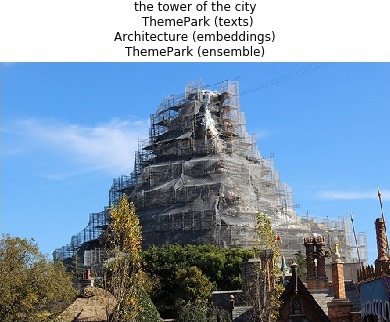}
 \caption{}
\end{subfigure}
\begin{subfigure}{0.49\textwidth}
 \centering
 \includegraphics[height=7cm]{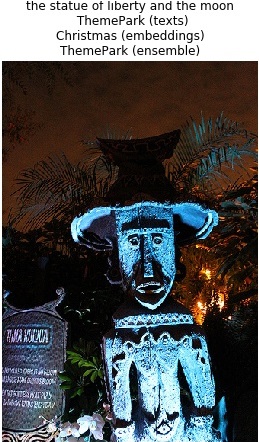}
 \caption{}
\end{subfigure}
 \caption{Sample results of event recognition }
\label{fig:event_recognition_sample}
\end{figure*}

In general we do not expect that classification of short textual descriptions is more accurate than the conventional image recognition methods. Nevertheless, we believe that the presence of image captions in an ensemble of classifiers can significantly improve its diversity. Moreover, as the captions are generated based on the extracted feature vector $\mathbf{x}$, only one inference in the CNN is required if we combine conventional general classifier of embeddings from pre-trained CNN and the image captions. In this paper the outputs of individual classifiers are combined in a simple voting with soft aggregation. In particular, we compute aggregated confidences as the weighted sum of outputs of individual classifier: 

\begin{equation}
\label{eq:ensemble}
\mathbf{p}_{ensemble}=[p_1,...,p_C]=w \cdot \mathbf{p}_{emb}+(1-w)\mathbf{p}_{txt}.
\end{equation}

The decision is taken in favor of the class with the maximal confidence:

\begin{equation}
\label{eq:final}
c^*=\underset{c \in \{1,...,C\}}\argmax p_c. 
\end{equation}

The weight $w \in [0,1]$ in (\ref{eq:ensemble}) can be chosen using a special validation subset in order to obtain the highest accuracy of criterion (\ref{eq:final}).
 
Let us provide qualitative examples for the usage of our pipeline (Fig.~\ref{cap_pipeline}). The results of (correct) event recognition using our ensemble are presented in Fig.~\ref{fig:event_recognition_sample}. Here the first line of the title contains the generated image caption. In addition, the title displays the result of event recognition using captions $\mathbf{t}$ (second line), embeddings $\mathbf{x}_{emb}$ (third line), and the whole ensemble (last line). As one can notice, the single classification of captions is not always correct. However, our ensemble is able to obtain reliable solution even when individual classifiers make wrong decisions.

\section{EXPERIMENTAL STUDY}\label{sec:4}

\subsection{Event recognition in a gallery of photos}

Only a limited number of datasets is available for event recognition in personal photo-collections~\cite{ahmad2019deep}. Hence, we examined two main datasets in this field, namely:

\begin{enumerate}
\item PEC~\cite{bossard2013event} with 61,364 images from 807 collections of 14 social event classes (birthday, wedding, graduation, etc.). We used its split provided by authors: the training set with 667 albums (50,279 images) and testing set with 140 albums (11,085 images). 
\item ML-CUFED~\cite{wang2017recognizing} contains 23 common event types. Each album is associated with several events, i.e., it is a multi-label classification task. Conventional split into the training set (75,377 photos, 1507 albums) and test set (376 albums with 19,420 photos) was used. 
\end{enumerate}

The features were extracted using the scene recognition models (Inception v3 and MobileNet v2 with $\alpha=1$ and $\alpha=1.4$) pre-trained on the Places2 dataset~\cite{zhou2018places}. We used two techniques in order to obtain a final descriptor of a set of images, namely, 1) simple averaging of features of individual images in a set (AvgPool); and 2) our implementation of neural attention mechanism (\ref{eq:weighted_sum})-(\ref{eq:attention_1}) for $L_2$-normed features. In the former case the linear SVM classifier from scikit-learn library was used as $\mathcal{C}$, because it has higher accuracy than RF, k-NN and RBF SVM. In the latter case the weights of the attention-based network (Fig.~\ref{fig:network}) are learned using the sets with $S=10$ randomly chosen images from all albums in order to make identical shape of input tensors. As a result, 667 training subsets and 1507 subsets with $S=10$ images were obtained for PEC and ML-CUFED, respectively. As the ML-CUFED contains multiple labels per each album, we use sigmoid activations and binary cross-entropy loss. Conventional Softmax activations and categorical cross-entropy are applied for the PEC. The model was learned using ADAM optimizer (learning rate 0.001) for 10 epochs with early stop in Keras 2.3 framework with TensorFlow 1.15 backend. 

\begin{table}
\begin{center}
{\caption{Accuracy (\%) of event recognition in a set of images (album).}\label{table_attention_album}}
\begin{tabular}{lp{0.4\linewidth}cc}
\hline
\rule{0pt}{12pt}
CNN & Aggregation & PEC & ML-CUFED
\\
\hline
\\[-6pt]
MobileNet2, & AvgPool & 86.42 & 81.38 \\
$\alpha=1.0$  & Attention & 89.29 & 84.04 \\ \hline
MobileNet2, & AvgPool & 87.14 & 81.91 \\
$\alpha=1.4$  & Attention & 87.36 & 84.31 \\ \hline
Inception & AvgPool & 86.43 & 82.45 \\
v3 & Attention & 87.86 & 84.84 \\ \hline
& CNN-LSTM-Iterative~\cite{wang2017recognizing} & 84.5 & 79.3 \\
AlexNet & Aggregation of representative features~\cite{wu2015learning} & 87.9 & 84.5 \\ \hline
& CNN-LSTM-Iterative~\cite{wang2017recognizing}& 84.5 & 71.7 \\
ResNet-101 & Aggregation of representative features~\cite{wu2015learning} & 89.1 & 83.4 \\
\hline
\\[-6pt]
\end{tabular}
\end{center}
\end{table}

The recognition accuracies of the pre-trained CNN are presented in Table~\ref{table_attention_album}. Here we computed the multi-label accuracy for ML-CUFED so that prediction is assumed to be correct if it corresponds to any label associated with an album. In this table we provided the best-known results for these datasets~\cite{wang2017recognizing,wu2015learning}.

Here in all cases the attention-based aggregation is 1-3\% more accurate when compared to classification of average features. As one can notice, the proposed implementation of attention mechanism achieves the state-of-the-art results, though we used much faster CNNs (MobileNet and Inception rather than AlexNet and ResNet-101) and do not consider sequential nature of photos in an album in our attention-based network (Fig.~\ref{fig:network}). The most remarkable fact here is that the best results for the PEC are achieved for the most simple model (MobileNet v2, $\alpha=1.0$), which can be explained by the lack of training data for this particular dataset.

As we claimed above, in general there is no information about albums in a gallery. Hence, event should be assigned to all photos individually. In the next experiment we directly assigned the collection-level first label to each image contained in both datasets and simply use the image itself for event recognition, without any meta information. In addition to baseline approach (Subsection~\ref{sec:3.1}) we used hierarchical agglomerative clustering of entire testing gallery. We report only the best results achieved by the average linkage clustering of embeddings $\mathbf{x}_t$ extracted by pre-trained CNN and confidence scores $\mathbf{p}_t$. In the former case we used both Euclidean ($L_2$) and chi-squared ($\chi^2$) distances. As the confidence scores returned by decision\_function for LinearSVC are not always non-negative, only Euclidean distance is implemented for the confidence scores. The results are shown in Table~\ref{table_single}.

\begin{table}
\begin{center}
{\caption{Accuracy (\%) of event recognition in a single image.}\label{table_single}}
\begin{tabular}{lccccc}
\hline
\rule{0pt}{12pt}
Dataset & CNN & Baseline & \multicolumn{2}{c}{Embeddings} & Scores \\
& & & $L_2$ & $\chi^2$ & $L_2$ \\
\hline
\\[-6pt]
 & MobileNet2, $\alpha=1.0$ & 58.32 & 60.42 & 60.69 & 58.44 \\
PEC & MobileNet2, $\alpha=1.4$ & 60.34 & 61.25 & 61.92 & 60.58 \\
 & Inception v3 & 61.82 & 64.19 & 64.22 & 61.97 \\ \hline
ML- & MobileNet2, $\alpha=1.0$ & 54.41 & 57.03 & 57.45 & 54.56 \\
CUFED & MobileNet2, $\alpha=1.4$ & 53.54 & 54.97 & 55.98 & 54.03 \\
 & Inception v3 & 57.26 & 59.19 & 60.12 & 57.87\\
\hline
\\[-6pt]
\end{tabular}
\end{center}
\end{table}

\begin{table*}[t!]
\begin{center}
{\caption{Accuracy (\%) of the proposed approach, PEC.}\label{table_combined_pec}}
\begin{tabular}{lccccccc}
\hline
\rule{0pt}{12pt}
CNN & Aggregation & Baseline & \multicolumn{2}{c}{Embeddings} & \multicolumn{2}{c}{Scores}& Scores ($L_2$-normed) \\
 & & & $L_2$ & $\chi^2$ & $L_2$ & $\chi^2$ & $L_2$
\\
 \hline
\\[-6pt]
MobileNet2, $\alpha=1.0$ & AvgPool & 58.32 & $66.85 \pm 0.59$ & $68.52 \pm 0.89$ & $71.08 \pm 0.59$ & - & $72.68 \pm 0.56$ \\
(pre-trained), embeddings & Attention & $54.43$ & $68.51 \pm 0.41$ & $70.65 \pm 1.20$ & $74.49 \pm 0.70$ & - & $80.48 \pm 1.01$ \\ \hline
MobileNet2, $\alpha=1.4$& AvgPool & $60.34$ & $68.85 \pm 0.59$ & $69.57 \pm 0.57$ & $72.59 \pm 1.49$ & - & $73.49 \pm 0.86$ \\
(pre-trained), embeddings & Attention & $55.36$ & $70.53 \pm 0.79$ & $71.16 \pm 0.72$ & $78.20 \pm 1.47$ & - & $81.27 \pm 0.81$ \\ \hline
MobileNet2, $\alpha=1.4$& AvgPool & $61.89$ & - & - & $75.66 \pm 0.55$ & $76.96 \pm 0.97$ & - \\
(fine-tuned), scores & Attention & $61.55$ & - & - & $78.77 \pm 0.49$ & $81.33 \pm 0.69$ & - \\ \hline
Inception v3& AvgPool & $61.82$ & $72.29 \pm 1.28$ & $72.32 \pm 1.54$ & $74.54 \pm 1.04$ & - & $76.48 \pm 0.47$ \\
(pre-trained), embeddings & Attention & $56.94$ & $72.38 \pm 1.13$ & $71.96 \pm 0.67$ & $76.76 \pm 0.70$ & - & $80.17 \pm 1.14$ \\ \hline
Inception v3& AvgPool & $63.56$ & - & - & $78.87 \pm 0.67$ & $79.92 \pm 0.65$ & -\\
(fine-tuned), scores & Attention & $62.91$ & - & - & $81.03 \pm 0.77$ & $81.95 \pm 1.11$ & -\\
\hline
\\[-6pt]
\end{tabular}
\end{center}
\end{table*}

\begin{table*}[t!]
\begin{center}
{\caption{Accuracy (\%) of the proposed approach, ML-CUFED.}\label{table_combined_cufed}}
\begin{tabular}{lccccccc}
\hline
\rule{0pt}{12pt}
CNN & Aggregation & Baseline & \multicolumn{2}{c}{Embeddings} & \multicolumn{2}{c}{Scores}& Scores ($L_2$-normed) \\
 & & & $L_2$ & $\chi^2$ & $L_2$ & $\chi^2$ & $L_2$
\\
 \hline
\\[-6pt]
MobileNet2, $\alpha=1.0$ & AvgPool & $54.41$ & $67.54 \pm 0.76$ & $67.42 \pm 0.93$ & $69.83 \pm 0.74$ & - & $70.42 \pm 0.41$ \\
(pre-trained), embeddings & Attention & $51.05$ & $68.71 \pm 0.71$ & $68.55 \pm 0.61$ & $71.44 \pm 0.82$ & - & $71.61 \pm 0.69$ \\ \hline
MobileNet2, $\alpha=1.4$& AvgPool & $53.54$ & $66.93 \pm 0.60$ & $67.21 \pm 0.55$ & $68.56 \pm 0.73$ & - & $69.47 \pm 0.36$ \\
(pre-trained), embeddings & Attention & $51.12$ & $68.34 \pm 0.68$ & $68.62 \pm 0.50$ & $70.79 \pm 0.75$ & - & $71.78 \pm 0.74$ \\ \hline
MobileNet2, $\alpha=1.4$& AvgPool & $56.01$ & - & - & $70.57 \pm 0.48$ & $71.61 \pm 0.28$ & - \\
(fine-tuned), scores & Attention & $56.09$ & - & - & $72.90 \pm 0.59$ & $73.46 \pm 0.58$ & - \\ \hline
Inception v3& AvgPool & $57.26$ & $69.91 \pm 0.58$ & $70.01 \pm 0.62$ & $72.25 \pm 0.61$ & - & $72.78 \pm 0.71$ \\
(pre-trained), embeddings & Attention & $50.89$ & $69.30 \pm 0.47$ & $68.52 \pm 0.89$ & $72.73 \pm 0.72$ & - & $73.00 \pm 0.65$ \\ \hline
Inception v3& AvgPool & $57.12$ & - & - & $72.18 \pm 0.63$ & $73.20 \pm 0.74$ & - \\
(fine-tuned), scores & Attention & $57.29$ & - & - & $73.06 \pm 0.74$ & $73.92 \pm 0.81$ & - \\
\hline
\\[-6pt]
\end{tabular}
\end{center}
\end{table*}

Here, firstly, the accuracy of event recognition in single images is 25-30\% lower than the accuracy of the album-based classification (Table~\ref{table_attention_album}). Secondly, clustering of the confidence scores at the output of the best classifier does not significantly influence the overall accuracy. Thirdly, hierarchical clustering with the chi-squared distance leads to slightly more accurate results than conventional Euclidean metric. Finally, preliminarily clustering of embeddings decreases the error rate of the baseline in only 1.2-2\% even if the distance threshold in clustering is carefully chosen. 

Let us demonstrate how the assumption about sequentially ordered photos in an album can increase the accuracy of event recognition. In order to make the task more complex, the following transformation of the order of testing photos was performed 10 times. We randomly shuffled the sequence of albums, and the photos in each album are also shuffled. In addition to the matching of confidences from decision\_function of the linearSVC we perform their $L_2$ normalization. Moreover, we fine-tuned CNNs using the unfolded training set $\mathbf{X}$ as follows. At first, the weights in the base part of the CNN were frozen and the new head (fully connected layer with $C$ outputs and Softmax activation) was learned during 10 epochs. Next, the weights in the whole CNN were learned during 3 epochs with 10-times lower learning rate.

The results (mean accuracy $\pm$ its standard deviation) of the proposed Algorithms~\ref{algorithm1},~\ref{algorithm2} for the PEC and the ML-CUFED are presented in Table~\ref{table_combined_pec} and Table~\ref{table_combined_cufed}, respectively. Here the attention mechanism provides up to 8\% lower error rates in most cases. It is remarkable that the matching of distances between $L_2$-normed confidences significantly improves the overall accuracy of attention model for the PEC (Table~\ref{table_combined_pec}), though our experiments did not show any improvements in conventional clustering from the previous experiment (Table~\ref{table_single}). The fine-tuned CNNs obviously lead to the most accurate decision, but the difference (0.1-1.6\%) with the best results of the pre-trained models is rather small. However, the latter do not require additional inference in existing scene recognition models, so the implementation of event recognition in an album will be very fast if the scenes should be additionally classified, e.g., for more detailed user modeling~\cite{savchenko2019user}. Surprisingly, computing the distance between confidence scores of classifiers ($\rho(\mathbf{p}_t,\mathbf{p}_{t-1})$) reduces the error rate of conventional matching of embeddings ($\rho(\mathbf{x}_t,\mathbf{x}_{t-1})$) on 2-7\%. Let us recall that conventional clustering of embeddings was 1-2\% more accurate when compared to the classifier's scores (Table~\ref{table_single}). It seems that the threshold $\rho_0$ can be estimated (Algorithm~\ref{algorithm2}) more reliably in this particular case when most images from the same event are matched in the prediction procedure (Algorithm~\ref{algorithm1}). Finally, the most important conclusion is that the proposed approach has 9-20\% higher accuracies when compared to baseline. Moreover, our algorithm is 13-16\% more accurate than classification of groups of photos obtained with hierarchical clustering (Table~\ref{table_single}).

\subsection{Event recognition in single photos}
In addition to PEC and ML-CUFED we examined WIDER (Web Image Dataset for Event Recognition)~\cite{xiong2015recognize} with 50,574 images and $C=61$ event categories (parade, dancing, meeting, press conference, etc). We used standard train/test split for all datasets proposed by their creators. In PEC and ML-CUFED the collection-level label is directly assigned to each image contained in this collection. We completely ignore any metadata, e.g., temporal information,  except the image itself similarly to the paper~\cite{wang2018transferring}.

As we mainly focus on possibility to implement offline event recognition on mobile devices~\cite{savchenko2019user}, in order to compare the proposed approach with conventional classifiers, we used MobileNet v2 with $\alpha=1$~\cite{cap_mobilenet} and Inception v4~\cite{cap_inc4} CNNs. At first, we pre-trained them on the Places2 dataset~\cite{zhou2018places} for feature extraction. The linear SVM classifier from scikit-learn library was used, because it has higher accuracy than other classifiers from this library (RF, k-NN and RBF SVM). Moreover, we fine-tuned these CNNs using the given training set as follows. At first, the weights in the base part of the CNN were frozen and the new head (fully connected layer with $C$ outputs and Softmax activation) was learned using ADAM optimizer (learning rate 0.001) for 10 epochs with early stop in Keras 2.2 framework with TensorFlow 1.15 backend. Next, the weights in the whole CNN were learned during 5 epochs using ADAM. Finally, the CNN was trained using SGD during 3 epochs with 10-times lower learning rate.

In addition, we used features from object detection models that are typical for event recognition~\cite{xiong2015recognize,savchenko2019user}. As many photos from the same event sometimes contains identical objects (e.g., ball in the football), they can be detected by contemporary CNN-based methods, i.e., SSDLite~\cite{cap_mobilenet} or Faster R-CNN~\cite{ren2015faster}. These methods detect the positions of several objects in the input image and predict the scores of each class from the predefined set of $K>1$ types. We extract the sparse $K$-dimensional vector of scores for each type of object. If there are several objects of the same type, the maximal score is stored in this feature vector~\cite{rassadin2019scene}. These feature vector is either classified by the linear SVM or used to train a feed-forward neural network with two hidden layers containing 32 units. Both classifiers were learned using the training set from each event dataset. In this study we examined SSD with MobileNet backbone and Faster R-CNN with InceptionResNet backbones. The models pre-trained on the Open Images Dataset v4 ($K=601$ objects) were taken from the TensorFlow Object Detection Model Zoo. 

Our preliminarily experimental study with the pre-trained image captioning models discussed in Section~\ref{sec:2} demonstrated that the best quality for MS COCO captioning dataset is achieved by the ARNet model~\cite{cap_ARNet}. Thus, in this experiment we used ARNet's encoder-decoder model.  However, it can be replaced to any other image captioning technique without modification of our event recognition algorithm. The ARNet was trained on the Conceptual Captions Dataset that contains more than 3.3M image-URL and caption pairs in the training set, and about 15 thousand pairs in the validation set. The feature extraction in encoder is implemented not only with he same CNNs (Inception and MobileNet v2). We extracted $|\tilde{V}|=5000$ most frequent words except special tokens $<START>$ and $<END>$. They are classified by either linear SVM or a feed-forward neural network with the same architecture as for object detection case. Again, these classifiers are trained from scratch given each event training set. The weight $w$ in our ensemble (Eq. 1) was estimated using the same set.

The results of the lightweight mobile (MobileNet and SSD object detector) and deep models (Inception and Faster R-CNN) for PEC, WIDER and ML-CUFED are presented in Tables~\ref{cap_pec},~\ref{cap_wider},~\ref{cap_cufed}, respectively. Here we added the best known results for the same experimental setups.

\begin{table}
\begin{center}
\caption{Event recognition accuracy (\%), PEC}
\label{cap_pec}
\begin{tabular}{cp{0.4\linewidth}p{0.15\linewidth}p{0.1\linewidth}}
\hline
\rule{0pt}{12pt}
Classifier & Features & Lightweight models & Deep models \\
\\
\hline
\\[-6pt]
\multirow{4}{*}{SVM} & Embeddings & 59.72 & 61.82 \\
 & Objects & 42.18 & 47.83 \\
 & Texts & 43.77 & 47.24 \\
& Proposed ensemble (\ref{eq:ensemble}), (\ref{eq:final}) & 60.56 & 62.87 \\
\hline
\multirow{4}{*}{Fine-tuned CNN} & Embeddings & 62.33 & 63.56 \\
 & Objects & 40.17 & 47.42 \\
 & Texts & 43.52 & 46.89 \\
& Proposed ensemble (\ref{eq:ensemble}), (\ref{eq:final}) & 63.38 & 65.12 \\
\hline
  \multicolumn{3}{c}{Aggregated SVM~\cite{bossard2013event}} & 41.4 \\
  \multicolumn{3}{c}{Bag of Sub-events~\cite{bossard2013event}} & 51.4 \\
  \multicolumn{3}{c}{SHMM~\cite{bossard2013event}} & 55.7 \\
\multicolumn{3}{c}{Initialization-based transfer learning~\cite{wang2018transferring}} & 60.6 \\
\multicolumn{3}{c}{Transfer learning of data and knowledge~\cite{wang2018transferring}} & 62.2 \\
\hline
\\[-6pt]
\end{tabular}
\end{center}
\end{table}

\begin{table}
\begin{center}
\caption{Event recognition accuracy (\%), WIDER}
\label{cap_wider}
\begin{tabular}{cp{0.4\linewidth}p{0.15\linewidth}p{0.1\linewidth}}
\hline
\rule{0pt}{12pt}
Classifier & Features & Lightweight models & Deep models \\
\\
\hline
\\[-6pt]
\multirow{4}{*}{SVM} & Embeddings & 48.31 & 50.48 \\
 & Objects & 19.91 & 28.66 \\
 & Texts & 26.38 & 31.89 \\
 & Proposed ensemble (\ref{eq:ensemble}), (\ref{eq:final})& 48.91 & 51.59 \\
\hline
\multirow{4}{*}{Fine-tuned CNN}  & Embeddings & 49.11 & 50.97 \\
 & Objects & 12.91 & 21.27 \\
 & Texts & 25.93 & 30.91 \\
& Proposed ensemble (\ref{eq:ensemble}), (\ref{eq:final})& 49.80 & 51.84 \\
\hline
\multicolumn{3}{c}{Baseline CNN~\cite{xiong2015recognize}} & 39.7 \\
\multicolumn{3}{c}{Deep channel fusion~\cite{xiong2015recognize}} & 42.4 \\
\multicolumn{3}{c}{Initialization-based transfer learning~\cite{wang2018transferring}} & 50.8 \\
\multicolumn{3}{c}{Transfer learning of data and knowledge~\cite{wang2018transferring}} & 53.0 \\
\hline
\\[-6pt]
\end{tabular}
\end{center}
\end{table}

\begin{table}
\begin{center}
\caption{Event recognition accuracy (\%), ML-CUFED}
\label{cap_cufed}
\begin{tabular}{cp{0.4\linewidth}p{0.15\linewidth}p{0.1\linewidth}}
\hline
\rule{0pt}{12pt}
Classifier & Features & Lightweight models & Deep models \\
\\
\hline
\\[-6pt]
\multirow{4}{*}{SVM} & Embeddings & 53.54 & 57.27 \\
 & Objects & 34.21 & 40.94 \\
 & Texts & 37.24 & 41.52 \\
 & Proposed ensemble (\ref{eq:ensemble}), (\ref{eq:final}) & 55.26 & 58.86 \\
\hline
 \multirow{4}{*}{Fine-tuned CNN} & Embeddings & 56.01 & 57.12 \\
 & Objects & 32.05 & 40.12 \\
 & Texts & 36.74 & 41.35 \\
& Proposed ensemble (\ref{eq:ensemble}), (\ref{eq:final})& 57.94 & 60.01 \\
\hline
\\[-6pt]
\end{tabular}
\end{center}
\end{table}

Certainly, the proposed recognition of image captions is not as accurate as conventional CNN-based features. However, classification of textual descriptions is much better than the random guess with accuracy $100\%/14 \approx 7.14\%$, $100\%/61 \approx 1.64\%$ and $100\%/23 \approx 4.35\%$ for PEC, WIDER and ML-CUFED, respectively. It is important to emphasize that our approach has lower error rate than classification of the features based on object detection in most case. This gain is especially noticeable for lightweight SSD models, which are 1.5-13\% less accurate than the proposed classification of image captions due to the limitations of SSD-based models to detect small objects (food, pets, fashion accessories, etc.). The Faster R-CNN-based detection features can be classified more accurately, but the inference in the Faster R-CNN with InceptionResNet backbone is several times slower than decoding in the ARNet (6-10 seconds vs 0.5-2 seconds on MacBook Pro 2015). 

Finally, the most appropriate way to use image captioning in event classification is its fusion with conventional CNNs. In such case we improved the previous state-of-the-art for PEC from 62.2\%~\cite{wang2018transferring} even for the lightweight models (63.38\%) if the fine-tuned CNNs are used in an ensemble. Our Inception-based model is even better (accuracy 65.12\%). We have not still reached the state-of-the-art accuracy 53\%~\cite{wang2018transferring} for the WIDER dataset, though our best accuracy (51.84\%) is up to 9\% higher when compared to the best results (42.4\%) from original paper~\cite{xiong2015recognize}. Our experimental setup for the ML-CUFED dataset is studied at first time here because this dataset is developed mostly for album-based event recognition.

In practice it is preferable to use pre-trained CNN as a feature extractor in order to prevent additional inference in fine-tuned CNN when it differs with the encoder in image captioning model. Unfortunately, the accuracies of SVM for pre-trained CNN features are 1.5-3\% lower when compared to the fine-tuned models for PEC and ML-CUFED. In this case additional inference may be acceptable. However, the difference in error rates between pre-trained and fine-tuned models for WIDER dataset is not significant, so that the pre-trained CNNs are definitely worth being used here.

\section{CONCLUSION}\label{sec:5}
We have shown that existing studies of event recognition cannot be directly used for processing of a gallery of mobile device because the albums of photos corresponding to the same event may be unavailable. The usage of event recognition in single images is possible but is very inaccurate even if similar photos are combined with a clustering (Table~\ref{table_single}). We have demonstrated that grouping of consecutive photos and attention-based recognition of resulted image sets (Algorithm~\ref{algorithm1}) can drastically improve the recognition accuracy (Tables~\ref{table_combined_pec},~\ref{table_combined_cufed}). It has been shown that the most important parameter, namely, distance threshold $\rho_0$, can be automatically estimated in our learning procedure (Algorithm~\ref{algorithm2}). It has been experimentally demonstrated that consecutive photos from the same album are better discovered if we match the confidence scores of classifier, which has been learned on unfolded training set $\mathbf{X}$.

In addition, we have proposed to apply generative models in classical discriminative task, namely, image captioning in event recognition in still images. We have presented the novel pipeline of visual preferences prediction using image captioning with classification of generated captions and retrieval of images based on their textual descriptions (Fig.~\ref{cap_pipeline}). It has been experimentally demonstrated that our approach is more accurate and faster than the widely-used image representations obtained by object detectors~\cite{xiong2015recognize,rassadin2019scene}. What is more important, generated caption provides additional diversity to conventional CNN-based recognition, which is especially useful for ensemble models.

Our engine has been implemented in the publicly available Android application (Fig.~\ref{fig:mobile_demo_UI}) that extracts the profile of user's interests. It is applicable for such personalized mobile services as recommender systems and target advertisements. 

The main disadvantage of the proposed approach is its lower accuracy (up to 8-11\%) when compared to the best models for the case of known borders of albums (Table~\ref{table_attention_album}). Moreover, short conceptual textual descriptions are obviously not enough to classify event categories with high accuracy even for a human due to errors and lack of specificity (see example of generated captions in Fig.~\ref{fig:event_recognition_sample}). Another disadvantage of the proposed approach is the need to repeat inference if fine-tuned CNN is applied in an ensemble. Hence, the decision-making time will be significantly increased though the overall accuracy becomes also higher in most cases (Tables~\ref{cap_pec},~\ref{cap_cufed}). 

Thus, in future it is possible to extend our algorithm by, e.g., replacing the pre-defined metric $\rho(\mathbf{p}_t,\mathbf{p}_{t-1})$ to a metric learned on a given training set~\cite{goodfellow2016deep}. Secondly, our attention model does not work well for single photos: its accuracy for the baseline with pre-trained CNNs is 4-5\% worth than the accuracy of linear SVM (row ``AvgPool" in Tables~\ref{table_combined_pec},~\ref{table_combined_cufed}). Hence, it is desirable to examine appropriate enhancements of attention model that are suitable even for small input set~\cite{guo2017multigranular,wu2015learning}. Finally, it is necessary to make classification of generated captions more accurate. Though our preliminary experiments of LSTMs did not decrease the error rate of our simple approach with linear SVM and one-hot encoded words, we strongly believe that a thorough study of the RNN-based classifiers of generated textual descriptors is required. 

\ack This research is based on the work supported by Samsung Research, Samsung Electronics.

\bibliography{for_arxiv}

\begin{thebibliography}{10}

\bibitem{ahmad2019deep}
Kashif Ahmad and Nicola Conci, `How deep features have improved event
  recognition in multimedia: A survey', {\em ACM Transactions on Multimedia
  Computing, Communications, and Applications (TOMM)}, {\bf 15}(2), ~39,
  (2019).

\bibitem{ahmad2017event}
Kashif Ahmad, Nicola Conci, Giulia Boato, and Francesco~GB De~Natale, `Event
  recognition in personal photo collections via multiple instance
  learning-based classification of multiple images', {\em Journal of Electronic
  Imaging}, {\bf 26}(6),  060502, (2017).

\bibitem{ashour2019improve}
Wesam~M Ashour, Riham~Z Muqat, Alaaeddin~B AlQazzaz, and Saeb~R AbdElnabi,
  `Improve basic sequential algorithm scheme using ant colony algorithm', in
  {\em Proceedings of the 7th Palestinian International Conference on
  Electrical and Computer Engineering (PICECE)}, pp. 1--6. IEEE, (2019).

\bibitem{bacha2016event}
Siham Bacha, Mohand~Said Allili, and Nadjia Benblidia, `Event recognition in
  photo albums using probabilistic graphical models and feature relevance',
  {\em Journal of Visual Communication and Image Representation}, {\bf 40},
  546--558, (2016).

\bibitem{bossard2013event}
Lukas Bossard, Matthieu Guillaumin, and Luc Van~Gool, `Event recognition in
  photo collections with a stopwatch {HMM}', in {\em Proceedings of the
  International Conference on Computer Vision (ICCV)}, pp. 1193--1200. IEEE,
  (2013).

\bibitem{cap_ARNet}
Xinpeng Chen, Lin Ma, Wenhao Jiang, Jian Yao, and Wei Liu, `Regularizing rnns
  for caption generation by reconstructing the past with the present', in {\em
  Proceedings of the IEEE Conference on Computer Vision and Pattern Recognition
  (CVPR)}, (2018).

\bibitem{francois2017deep}
Francois Chollet, {\em Deep learning with Python}, Manning Publications
  Company, 2017.

\bibitem{dao2011signature}
Minh-Son Dao, Duc-Tien Dang-Nguyen, and Francesco~GB De~Natale,
  `Signature-image-based event analysis for personal photo albums', in {\em
  Proceedings of the 19th International Conference on Multimedia (ACM MM)}, pp.
  1481--1484. ACM, (2011).

\bibitem{escalera2015chalearn}
Sergio Escalera, Junior Fabian, Pablo Pardo, Xavier Baro, Jordi Gonzalez,
  Hugo~J Escalante, Dusan Misevic, Ulrich Steiner, and Isabelle Guyon,
  `Chalearn looking at people 2015: Apparent age and cultural event recognition
  datasets and results', in {\em Proceedings of the International Conference on
  Computer Vision Workshops (ICCVW)}, pp. 1--9, (2015).

\bibitem{geng2018classifying}
Ming Geng, Yukun Li, and Fenglian Liu, `Classifying personal photo collections:
  an event-based approach', in {\em Proceedings of the Asia-Pacific Web (APWeb)
  and Web-Age Information Management (WAIM) Joint International Conference on
  Web and Big Data}, pp. 201--215. Springer, (2018).

\bibitem{goodfellow2016deep}
Ian Goodfellow, Yoshua Bengio, and Aaron Courville, {\em Deep learning}, MIT
  Press, 2016.

\bibitem{grechikhin2019user}
Ivan Grechikhin and Andrey~V Savchenko, `User modeling on mobile device based
  on facial clustering and object detection in photos and videos', in {\em
  Proceedings of the Iberian Conference on Pattern Recognition and Image
  Analysis}, pp. 429--440. Springer, (2019).

\bibitem{guo2017multigranular}
Cong Guo, Xinmei Tian, and Tao Mei, `Multigranular event recognition of
  personal photo albums', {\em IEEE Transactions on Multimedia}, {\bf 20}(7),
  1837--1847, (2017).

\bibitem{hossain2019comprehensive}
MD~Hossain, Ferdous Sohel, Mohd~Fairuz Shiratuddin, and Hamid Laga, `A
  comprehensive survey of deep learning for image captioning', {\em ACM
  Computing Surveys (CSUR)}, {\bf 51}(6),  118, (2019).

\bibitem{howard2017MobileNets}
Andrew~G Howard, Menglong Zhu, Bo~Chen, Dmitry Kalenichenko, Weijun Wang,
  Tobias Weyand, Marco Andreetto, and Hartwig Adam, `{MobileNets}: Efficient
  convolutional neural networks for mobile vision applications', {\em arXiv
  preprint arXiv:1704.04861}, (2017).

\bibitem{kuzovkin2019context}
Dmitry Kuzovkin, Tania Pouli, Olivier~Le Meur, R{\'e}mi Cozot, Jonathan Kervec,
  and Kadi Bouatouch, `Context in photo albums: Understanding and modeling user
  behavior in clustering and selection', {\em ACM Transactions on Applied
  Perception (TAP)}, {\bf 16}(2), ~11, (2019).

\bibitem{lonn2019smartphone}
Stefan Lonn, Petia Radeva, and Mariella Dimiccoli, `Smartphone picture
  organization: A hierarchical approach', {\em Computer Vision and Image
  Understanding}, {\bf 187},  102789, (2019).

\bibitem{cap_NBT}
Jiasen Lu, Jianwei Yang, Dhruv Batra, and Devi Parikh, `Neural baby talk', in
  {\em Proceedings of the IEEE Conference on Computer Vision and Pattern
  Recognition (CVPR)}, (2018).

\bibitem{cap_mRNN}
Junhua Mao, Wei Xu, Yi~Yang, Jiang Wang, and Alan~L. Yuille, `Deep captioning
  with multimodal recurrent neural networks ({m-RNN})', in {\em Proceedings of
  the International Conference on Learning Representations (ICLR)}, (2015).

\bibitem{rassadin2019scene}
Alexandr Rassadin and Andrey Savchenko, `Scene recognition in user preference
  prediction based on classification of deep embeddings and object detection',
  in {\em Proceedings of the International Symposium on Neural Networks
  (ISNN)}, volume 11555, pp. 422--430. Springer, (2019).

\bibitem{ren2015faster}
Shaoqing Ren, Kaiming He, Ross Girshick, and Jian Sun, `Faster {R-CNN}: Towards
  real-time object detection with region proposal networks', in {\em Advances
  in Neural Information Processing Systems (NIPS)}, pp. 91--99, (2015).

\bibitem{sandler_inverted_2018}
Mark Sandler, Andrew Howard, Menglong Zhu, Andrey Zhmoginov, and Liang-Chieh
  Chen, `{MobilenetV2}: Inverted residuals and linear bottlenecks', in {\em
  Proceedings of the International Conference on Computer Vision and Pattern
  Recognition (CVPR)}, pp. 4510--4520, (2018).

\bibitem{cap_mobilenet}
Mark Sandler, Andrew~G. Howard, Menglong Zhu, Andrey Zhmoginov, and
  Liang{-}Chieh Chen, `Mobilenetv2: Inverted residuals and linear bottlenecks',
  in {\em Proceedings of the IEEE Conference on Computer Vision and Pattern
  Recognition, {CVPR} 2018, Salt Lake City, UT, USA, June 18-22, 2018}, pp.
  4510--4520, (2018).

\bibitem{savchenko2019efficient}
Andrey~V. Savchenko, `Efficient facial representations for age, gender and
  identity recognition in organizing photo albums using multi-output
  {ConvNet}', {\em PeerJ Computer Science}, {\bf 5}(e197), (2019).

\bibitem{savchenko2019sequential}
Andrey~V. Savchenko, `Sequential three-way decisions in multi-category image
  recognition with deep features based on distance factor', {\em Information
  Sciences}, {\bf 489},  18--36, (2019).

\bibitem{savchenko2019user}
Andrey~V Savchenko, Kirill~V Demochkin, and Ivan~S Grechikhin, `User preference
  prediction in visual data on mobile devices', {\em arXiv preprint
  arXiv:1907.04519}, (2019).

\bibitem{sokolova2017organizing}
Anastasiia~D Sokolova, Angelina~S Kharchevnikova, and Andrey~V Savchenko,
  `Organizing multimedia data in video surveillance systems based on face
  verification with convolutional neural networks', in {\em Proceedings of the
  International Conference on Analysis of Images, Social Networks and Texts
  (AIST)}, pp. 223--230. Springer, (2017).

\bibitem{cap_inc4}
Christian Szegedy, Sergey Ioffe, Vincent Vanhoucke, and Alex~A. Alemi,
  `Inception-v4, {Inception-ResNet} and the impact of residual connections on
  learning', in {\em Proceedings of the International Conference on Learning
  Representations (ICLR) Workshop}, (2016).

\bibitem{vijayaraju2019image}
Nivetha Vijayaraju, `Image retrieval using image captioning', {\em Master's
  Projects}, {\bf 687}, (2019).

\bibitem{cap_ST}
O.~{Vinyals}, A.~{Toshev}, S.~{Bengio}, and D.~{Erhan}, `Show and tell: Lessons
  learned from the 2015 mscoco image captioning challenge', {\em IEEE
  Transactions on Pattern Analysis and Machine Intelligence}, {\bf 39}(4),
  652--663, (2017).

\bibitem{wang2018transferring}
Limin Wang, Zhe Wang, Yu~Qiao, and Luc Van~Gool, `Transferring deep object and
  scene representations for event recognition in still images', {\em
  International Journal of Computer Vision}, {\bf 126}(2-4),  390--409, (2018).

\bibitem{wang2017recognizing}
Yufei Wang, Zhe Lin, Xiaohui Shen, Radom{\'\i}r Mech, Gavin Miller, and
  Garrison~W Cottrell, `Recognizing and curating photo albums via
  event-specific image importance', in {\em Proceedings of British Conference
  on Machine Vision (BMVC)}, (2017).

\bibitem{wu2015learning}
Zifeng Wu, Yongzhen Huang, and Liang Wang, `Learning representative deep
  features for image set analysis', {\em IEEE Transactions on Multimedia}, {\bf
  17}(11),  1960--1968, (2015).

\bibitem{xiong2015recognize}
Yuanjun Xiong, Kai Zhu, Dahua Lin, and Xiaoou Tang, `Recognize complex events
  from static images by fusing deep channels', in {\em Proceedings of the
  International Conference on Computer Vision and Pattern Recognition (CVPR)},
  pp. 1600--1609, (2015).

\bibitem{cap_SAT}
Kelvin Xu, Jimmy~Lei Ba, Ryan Kiros, Kyunghyun Cho, Aaron Courville, Ruslan
  Salakhutdinov, Richard~S. Zemel, and Yoshua Bengio, `Show, attend and tell:
  Neural image caption generation with visual attention', in {\em Proceedings
  of the International Conference on International Conference on Machine
  Learning (ICML)}, pp. 2048--2057, (2015).

\bibitem{yang2017neural}
J.~Yang, P.~Ren, D.~Zhang, D.~Chen, F.~Wen, H.~Li, and G.~Hua, `Neural
  aggregation network for video face recognition', in {\em Proceedings of the
  International Conference on Computer Vision and Pattern Recognition (CVPR)},
  pp. 5216--5225. IEEE, (2017).

\bibitem{zhou2018places}
Bolei Zhou, Agata Lapedriza, Aditya Khosla, Aude Oliva, and Antonio Torralba,
  `Places: A 10 million image database for scene recognition', {\em IEEE
  Transactions on Pattern Analysis and Machine Intelligence}, {\bf 40}(6),
  1452--1464, (2018).

\end{thebibliography}
\end{document}